\documentclass{article}

\usepackage[preprint]{neurips_2025}

\makeatletter
\DeclareRobustCommand\onedot{\futurelet\@let@token\@onedot}
\def\@onedot{\ifx\@let@token.\else.\null\fi\xspace}

\def\eg{\emph{e.g}\onedot,~} 
\def\ie{\emph{i.e}\onedot,~}

\makeatother

\usepackage[utf8]{inputenc} % allow utf-8 input
\usepackage[T1]{fontenc}    % use 8-bit T1 fonts
\usepackage{url}            % simple URL typesetting
\usepackage{booktabs}       % professional-quality tables
\usepackage{amsfonts}       % blackboard math symbols
\usepackage{nicefrac}       % compact symbols for 1/2, etc.
\usepackage{microtype}      % microtypography
\usepackage{xcolor}         % colors

\definecolor{refblue}{rgb}{0.21,0.49,0.74}
\usepackage[pagebackref,breaklinks,colorlinks,allcolors=refblue]{hyperref}

\usepackage{graphicx}	
\usepackage{amsmath}	
\usepackage{amssymb}
\usepackage{epsfig}
\usepackage{caption}
\usepackage{float}
\usepackage{placeins}
\usepackage{color, colortbl}
\usepackage{stfloats}
\usepackage{enumitem}
\usepackage{tabularx}
\usepackage{xstring}
\usepackage{multirow}
\usepackage{xspace}
\usepackage{subcaption}
\usepackage[hang,flushmargin]{footmisc}
\usepackage{duckuments}
\usepackage{lipsum}
\usepackage{wrapfig}

\usepackage[capitalize]{cleveref}
\crefname{section}{Section}{Sections}
\crefname{table}{Table}{Tables}
\crefname{figure}{Figure}{Figures}
\crefname{equation}{Equation}{Equations}
\crefname{appendix}{Appendix}{Appendices}

\newcommand{\ours}{MMP}
\newcommand{\duster}{DUSt3R}
\newcommand{\monster}{MonST3R}

\title{Learning Multi-frame and Monocular Prior for Estimating Geometry in Dynamic Scenes}

\author{%
  Seong Hyeon Park \\
  KAIST \\
  \texttt{seonghyp@kaist.ac.kr} \\
  \And
  Jinwoo Shin \\
  KAIST and RLWRLD \\
  \texttt{jinwoos@kaist.ac.kr}
}

\begin{document}
\maketitle

\begin{abstract}
In monocular videos that capture dynamic scenes, estimating the 3D geometry of video contents has been a fundamental challenge in computer vision.
Specifically, the task is significantly challenged by the object motion, where existing models are limited to predict only partial attributes of the dynamic scenes, such as depth or pointmaps spanning only over a pair of frames.
Since these attributes are inherently noisy under multiple frames, test-time global optimizations are often employed to fully recover the geometry, which is liable to failure and incurs heavy inference costs.
To address the challenge, we present a new model, coined~\ours, to estimate the geometry in a feed-forward manner, which produces a dynamic pointmap representation that evolves over multiple frames.
Specifically, based on the recent Siamese architecture, we introduce a new trajectory encoding module to project point-wise dynamics on the representation for each frame, which can provide significantly improved expressiveness for dynamic scenes.
%
% Furthermore, we propose a feed-forward refinement technique, which enables referring to cached representation across different frame sets, enhancing the quality of pointmaps over a long prediction horizon.
%
% Our design ensures the compatibility with the Siamese architecture so that the model can leverage pre-training on extensive static scenes and pair-wise configurations, including the recent state-of-the-art.
%
In our experiments, we find~\ours~can achieve state-of-the-art quality in feed-forward pointmap prediction, \eg 15.1\% enhancement in the regression error.
\end{abstract}

\section{Introduction}
\label{sec:intro}

Understanding dynamic video scenes is a highly desirable ability for AI systems to thrive in the real world.
Specifically, the task of 4D geometry estimation has been a fundamental challenge in computer vision, which aims to reconstruct physical 3D shapes in a dynamic scene observed as monocular video frames \citep{mustafa2016temporally, kumar2017, barsan2018robust, luiten2020, li2023dynibar, monst3r}.

Historically, this task has been tackled via multi-stage and optimization-based approaches \citep{luiten2020, li2023dynibar}. They employ individual models to predict attributes such as matching and depth as the first stage, and subsequently obtain a geometry model by combining the attributes through per-scene optimization.
However, these approaches tend to be computationally heavy and does not generalize well due to errors accumulated in the first stage.

To address the problem, recent works have pursued feed-forward designs which predict the geometry directly from the observed video frames \citep{monst3r, charatan2024pixelsplat, chen2024mvsplat}.
Notably, models based on the Siamese architecture \citep{dust3r, leroy2024grounding} have set state-of-the-art, which produce dense predictions associated with every pixels of the given frames, representing the 3D pointcloud in a shared coordinate system, \eg one frame's view.
This representation, referred to as the pointmap, can disentangle the effect of camera motion from 3D shapes, and has shown to better generalize to dynamic scenes than prior art \citep{monst3r}.

\begin{figure}[tp]
    \centering
    \includegraphics[width=\linewidth]{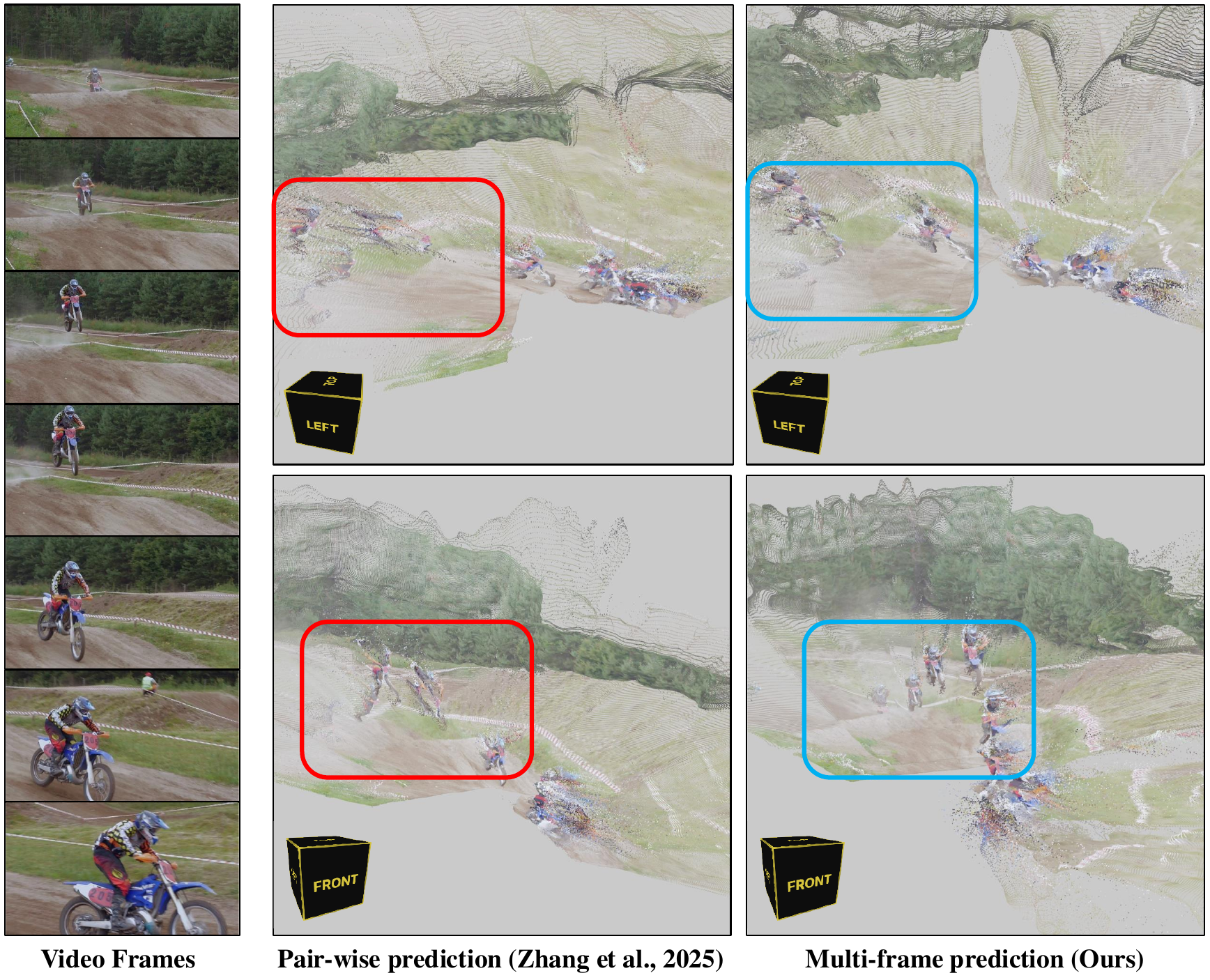}
    \caption{\textbf{Feed-forward pointmap prediction examples.} Given a set of 7 video frames from davis video dataset \citep{davis}, we visualize the corresponding pointmaps produced by a pair-wise baseline model \citep{monst3r} and our method in 2 different views (a top-left view in the upper row and a front-top view in the bottom row).
    While the pair-wise baseline suffers from inaccurate motion estimation in the pointmap (\eg the red boxes), our method can produce a pointmap that accurately represents dynamics over the frames (\eg the blue boxes).}
    \label{fig:problem_statement}
\end{figure}

However, the inherent drawback of the concurrent models is that they process only a pair of frames at once, and extending the number of frames is non-trivial in their Siamese architecture.
This poses significant limitations for processing complex dynamic scenes that require observing multiple frames beyond the pairs, and the models demonstrate sub-optimal performance, as depicted in \cref{fig:problem_statement}.
While existing methods mitigate the errors by accumulating pairwise estimates for multiple frames through the global optimization, they inevitably are computationally heavy and prone to errors, akin to the classical optimization-based approaches.

In this paper, we propose a new architecture which escalates the feed-forward 4D geometry estimation beyond the pair of frames.
Built on top of the Siamese design, our model adds only negligible amount of computation when compared to that of the global optimization in existing methods, but demonstrating up to 15.1$\%$ enhancement in the performance.
Specifically, we contribute the following new modules:
\begin{itemize}
\item \textbf{Trajectory Encoder} inserted to the Siamese transformer block to enable predicting dynamic pointmaps over multiple frames. This module significantly improves the expressiveness for dynamic scenes, yet ensures the compatibility with the existing pair-wise processing
\item \textbf{Feed-forward Refinement} given frame sets, which enables our model to refer to pointmap representation across inference iterations. We note that this module can save computations using a key-value caching technique. 

\end{itemize}
% Use figure* for multi-column figure

We provide the details of our method in \cref{sec:method}, the preliminaries of the Siamese architecture and our specific designs to address the problem. Then, we perform experiments benchmarking the quality of 4D geometry estimation in comparison with state-of-the-art baselines in \cref{sec:experiment}, where~\ours~achieves significant improvement in the feed-forward prediction quality.
\section{Related Work}
\label{sec:related}

\subsection{Static 3D geometry estimation}
Static 3D geometry estimation, or the 3D reconstruction, predicts 3D representation given a set of images, such as points and meshes \citep{qi2017pointnet, lin2018learning, wang2018pixel2mesh, gkioxari2019mesh}, voxels \citep{choy20163d, tulsiani2017multi, sitzmann2019deepvoxels}, or neural representations \citep{wang2021neus, peng2020convolutional, chen2019learning, wang2021ibrnet}.
Recently, DUSt3R \citep{dust3r} proposed the pointmap representation.
Given a pair of images, it predicts the pointcloud of every pixel in the images, in the coordinate system of one image's view point.
This new representation effectively disentangles the influence of camera motion and intrinsics from the 3D geometry, which has been shown to learn representation useful in downstream tasks. 

\subsection{4D geometry estimation}
Approaches for 4D geometry estimation of dynamic scenes split into optimization-based \citep{mustafa2016temporally, kumar2017, barsan2018robust, luiten2020, li2023dynibar} and feed-forward \citep{monst3r, charatan2024pixelsplat, chen2024mvsplat} models.
Due to a scarcity of training data for dynamic scenes, earlier approaches have focused on optimization-based models.
These methods, given video frames and attributes predicted by sub-task models (\eg optical flows \citep{teed2020raft, lipson2021raft}),
reconstruct the input video via test-time optimization of a 3D geometry representation \citep{mildenhall2021nerf, kerbl20233d}.
However, these approaches tend to be computationally heavy and do not generalize well due to errors accumulated in the pre-computed estimates.

Recently, feed-forward methods \citep{monst3r, charatan2024pixelsplat, chen2024mvsplat} have been proposed, which estimate 4D geometry directly from videos.
Specifically, \monster~\citep{monst3r} finds that the pointmap representation in \duster~\citep{dust3r} can be generalized to dynamic scenes by performing fine-tuning on dynamic 4D datasets.
However, as their architecture is still limited to pair-wise predictions, the quality of feed-forward tends to be sub-otpimal under complex dynamics.
Our work tackles this problem and enable a multi-frame processing for the pointmap prediction.

\section{Method}
\label{sec:method}
% Use figure* for multi-column figure
\begin{figure}[tp]
    \centering
    \includegraphics[width=\linewidth]{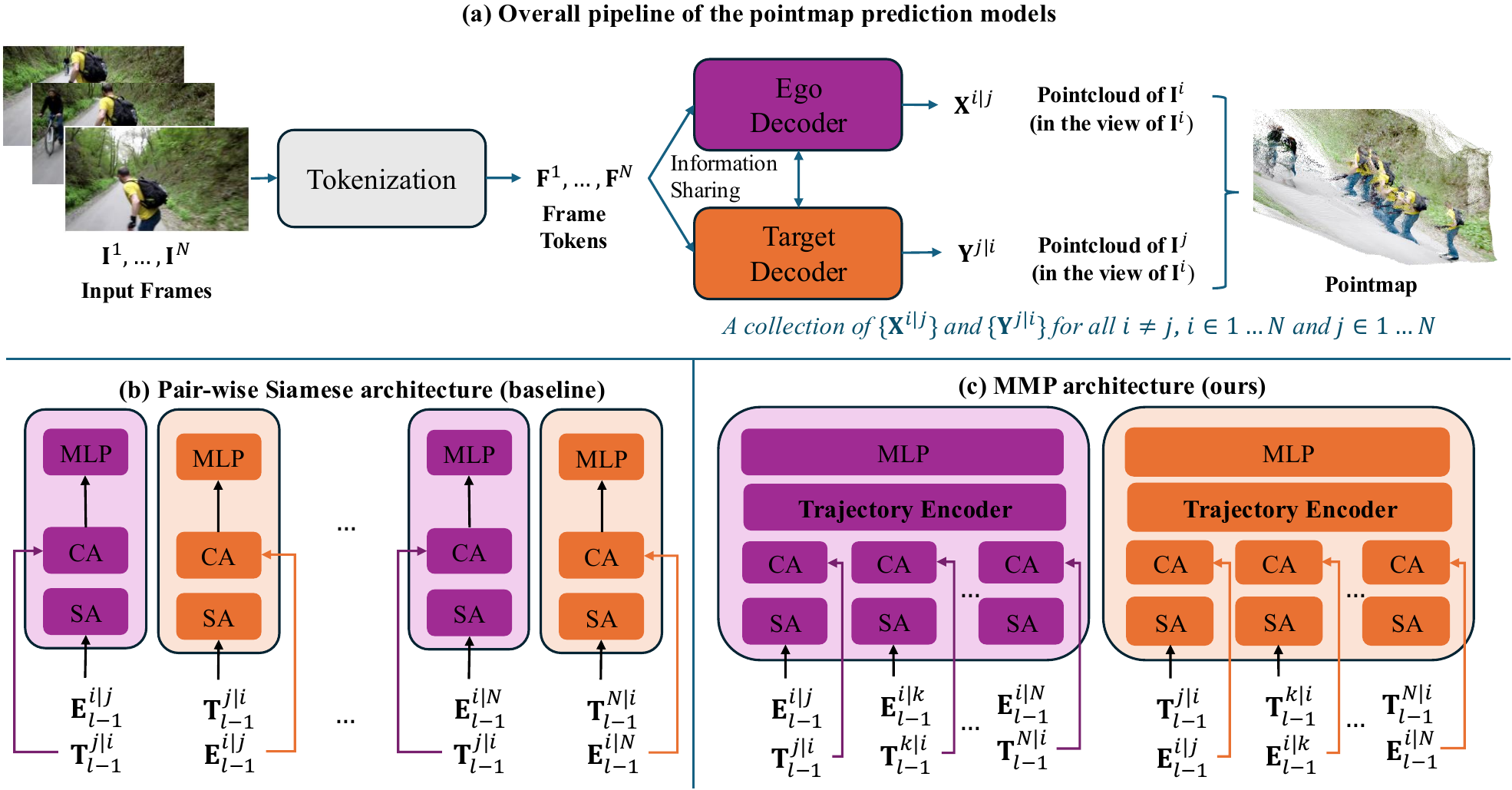}
    \caption{\textbf{Illustration of the prediction pipeline in~\ours.} The top figure (a) depeicts the overall pipeline of the pointmap prediction comprising the ego decoder (purple blcoks) and the target decoder (orange blocks), shared by both the Siamese baselines \citep{monst3r, dust3r} and our method. The bottom-left figure (b) illustrates the design of a decoder block in the baseline architecture, using the self-attention (SA) and the cross-attention (CA) mechanisms. The bottom-right figure (c) illustrates our architecture, equipped with the proposed trajectory encoder.}
    \label{fig:architecture}
\end{figure}

In this section, we provide the details of our architecture design for predicting pointmaps given a set of video frames. To begin, we review the baseline Siamese architecture in \cref{sec:method:siamese}, based on which we design a new architecture for our method.
Then, we introduce the trajectory encoder in \cref{sec:method:trajectory}, the key component of our method, which enables processing multiple frames beyond the limitation of the baseline.
Finally, in \cref{sec:method:refinement}, we describe the feed-forward refinement technique in our method.

As for the data notation, we denote scalars using normal letters, and tensors using bold letters with a superscript denoting frame indices. For example, an input RGB video frame is $\mathbf{I}^i\in\mathbb{R}^{U \times V \times 3}$, where $U \times V$ is the resolution, and a frame tokenization is $\mathbf{F}^{i}\in\mathbb{R}^{N \times D}$, where $N=\frac{U}{P} \times \frac{V}{P}$ with the patch size $P$ and the embedding dimension $D$. Tensors can be indexed, such as $\mathbf{F}^{i}(n)\in\mathbb{R}^{D}$, where $\mathbf{F}^{i} \equiv [ \mathbf{F}^{i}(1),...,\mathbf{F}^{i}(N) ]$. Finally, when emphasizing that a feature or data for frame $i$ is conditioned on the frame $j$, we use the superscript $i|j$, such as the pointmap output $\mathbf{X}^{i|j} \in\mathbb{R}^{U \times V \times 3} $, which we frequently use in \cref{sec:method:siamese}.

\subsection{Pair-wise Siamese architecture}
\label{sec:method:siamese}
Given a pair of frames $(\mathbf{I}^i, \mathbf{I}^j)$, the Siamese architecture aims to predict a pointmap: the ego pointcloud $\mathbf{X}^{i|j}$ which represents the 3D coordinate of $\mathbf{I}^i$, and the target pointcloud $\mathbf{Y}^{j|i}$ which represents the 3D coordinate of $\mathbf{I}^j$ following the camera view of $\mathbf{I}^i$, predicted by two separate decoders.
Specifically, concurrent models \citep{dust3r, monst3r} employ transformer blocks with relative position embedding as the decoder, which process the ego tokens $\mathbf{E}_{l}^{i|j} \in \mathbb{R}^{N \times D}$, and the target tokens $\mathbf{T}_{l}^{i|j} \in \mathbb{R}^{N \times D}$, where $l\in\{1,...,L\}$ is the transformer block index. The initial tokens ($l=0$) are ${\mathbf{F}^i :=\mathtt{Tokenization}(\mathbf{I}^i)}$, \ie $\mathbf{E}_{0}^{i|j} := \mathbf{F}^i$ and $\mathbf{T}_{0}^{j|i} := \mathbf{F}^j$.

In each transformer block (\cref{fig:architecture}b), the cross-attention $\mathtt{CA}(\cdot; \cdot)$, placed next to the self-attention $\mathtt{SA}(\cdot)$, conveys information between the ego and the target tokens, followed by the $\mathtt{MLP}(\cdot)$ layer producing the output of a block,
\begin{align}
\tilde{\mathbf{E}}_l^{i|j} &:= \mathtt{CA}\bigl( \mathtt{SA}(\mathbf{E}_{l-1}^{i|j}); \mathbf{T}_{l-1}^{j|i}\bigr) \label{eq:ego_ca} \\ 
\mathbf{E}_l^{i|j} &:=  \mathtt{MLP}\bigl(\tilde{\mathbf{E}}_l^{i|j}\bigr) \label{eq:ego_mlp} \\ 
\tilde{\mathbf{T}}_l^{j|i} &:= \mathtt{CA}\bigl(\mathtt{SA}(\mathbf{T}_{l-1}^{j|i});  \mathbf{E}_{l-1}^{i|j}\bigr) \label{eq:target_ca} \\ 
\mathbf{T}_l^{j|i} &:=  \mathtt{MLP}\bigl(\tilde{\mathbf{T}}_l^{j|i}\bigr), \label{eq:target_mlp}
\end{align}
assuming the skip-connections \citep{vaswani2017attention, he2016deep} existing in the layers.
To produce the output pointclouds, the DPT head layer \citep{dpt} is employed, which takes these block-wise tokens as the input, 
\begin{align}
\mathbf{X}^{i|j} &:= \mathtt{Head}\left(\mathbf{E}_0^{i|j}; \mathbf{E}_1^{i|j}; ...;\mathbf{E}_L^{i|j} \right) \label{eq:ego_head} \\ 
\mathbf{Y}^{j|i} &:= \mathtt{Head}\left(\mathbf{T}_0^{j|i}; \mathbf{T}_1^{j|i}; ...;\mathbf{T}_L^{j|i} \right). \label{eq:target_head}
\end{align}

Although we abuse the same notations $\mathtt{SA}$, $\mathtt{CA}$, $\mathtt{MLP}$, and $\mathtt{Head}$ for the two decoders and for all block indices $l\in\{1,...,L\}$, we note that their weight parameters are all different.

For most use cases, pair-wise models are executed twice, under the original and a swapped order of the input frames, producing $\{\mathbf{X}^{i|j}, \mathbf{X}^{j|i}, \mathbf{Y}^{j|i}, \mathbf{Y}^{i|j}\}$, which enables downstream tasks, such as 2-view geometry, estimating camera intrinsics and pose, etc.
When processing a greater number of frames $W>2$, inference is performed over all combinations, \eg for all $i \neq j$, $i \in \{1,...,W\}$ and  $j \in \{1,...,W\}$.
However, the pair-wise architecture is limited to process complex dynamic scenes, and the feed-forward performance is often sub-optimal, as we find in \cref{sec:experiment:feed_forward}.

\subsection{Trajectory encoder}
\label{sec:method:trajectory}
In this section, we describe our method to jointly process multiple frames (\ie $W>2$) to predict dynamic pointmaps.
To be specific, we enable it using the proposed trajectory encoder module, which collects the tokens in the same spatial index over the frames, then encode the inter-frame dynamics back to each token.

Without loss of generality, let us consider the frame $\mathbf{I}^W$, paired with others $\{ \mathbf{I}^1,...,\mathbf{I}^{W-1} \}$ and their corresponding tokens within the intermediate cross-attention stage of the decoder blocks in \cref{eq:ego_ca,eq:target_ca},
\begin{align}
% \text{set~}\tilde{\mathbf{E}}_{l}^{W|{\{k \neq j\}}}  &\equiv
\tilde{\mathbf{E}}_{l}^{W|{\{k<W\}}} &= \{ \tilde{\mathbf{E}}_{l}^{W|1},...,\tilde{\mathbf{E}}_{l}^{W|W-1} \} \label{eq:ego_tokenset} \\
% \text{set~}\tilde{\mathbf{T}}_{l}^{W|{\{k \neq j\}}} &\equiv
\tilde{\mathbf{T}}_{l}^{W|{\{k<W\}}} &= \{ \tilde{\mathbf{T}}_{l}^{W|1},...,\tilde{\mathbf{T}}_{l}^{W|W-1} \}. \label{eq:target_tokenset}
\end{align}
\vspace{1mm}

Intuitively, gathering from a same spatial index, \eg a stack of tokens $[ \tilde{\mathbf{T}}_{l}^{W|1}(n),...,\tilde{\mathbf{T}}_{l}^{W|W-1}(n) ]\in\mathbb{R}^{W \times D}$ by indexing each element in \cref{eq:ego_tokenset}, can represent the spatio-temporal dynamics of the patch region represented by $\mathbf{F}^W(n)$.
Therefore, projecting this feature onto each token of \cref{eq:ego_tokenset,eq:target_tokenset} can encode the dynamics.
Specifically, we apply an attention mechanism\footnote{We adjust the relative position embedding to encode a spatial index with the size $D/2$, and a time index with the size $D/2$.} with causal masks to implement the function, coined trajectory attention $\mathtt{TA}(\cdot; \cdot)$,

\begin{align}
\bar{\mathbf{E}}_{l}^{W|j} &:= \mathtt{TA}(\tilde{\mathbf{E}}_{l}^{W|j}; \tilde{\mathbf{E}}_{l}^{W|{\{k<W\}}}) \label{eq:ego_trajectory_naive} \\
\bar{\mathbf{T}}_{l}^{W|j} &:= \mathtt{TA}(\tilde{\mathbf{T}}_{l}^{W|j}; \tilde{\mathbf{T}}_{l}^{W|{\{k<W\}}}),\label{eq:target_trajectory_naive}
\end{align}
\vspace{1mm}
where
\vspace{1mm}
\begin{alignat}{3}
\bar{\mathbf{E}}_{l}^{W|j}(n) &= \mathtt{CA}\bigl(\tilde{\mathbf{E}}_{l}^{W|j}(n) ; [ \tilde{\mathbf{E}}_{l}^{W|1}(n),...,\tilde{\mathbf{E}}_{l}^{W|j}(n) ]\bigr) \\
\bar{\mathbf{T}}_{l}^{W|j}(n) &= \mathtt{CA}\bigl(\tilde{\mathbf{T}}_{l}^{W|j}(n) ; [ \tilde{\mathbf{T}}_{l}^{W|1}(n),...,\tilde{\mathbf{T}}_{l}^{W|j}(n) ]\bigr).
\end{alignat}
\vspace{1mm}

However, naively inserting this layer to each decoder block of a pre-trained Siamese model results in sub-optimal performance after training on dynamic scenes. 
In fact, prior art finds that retaining strong 3D prior learned from static datasets is crucial for learning 4D geometry \citep{monst3r}.
The trajectory attention deviate the computation graph of a pre-trained pair-wise model, losing the pre-trained 3D prior.
We note that it is also non-trivial to pre-train a multi-frame model from scratch, since the training data for 3D geometry is often a pair of images \citep{dust3r}, rather than a video stream data.

\vspace{2mm}
To address the problem, we aim to minimize the effect of modification in the initial state of the model. Specifically, inspired by model inflation techniques in video transformers \citep{timesformer, motionformer}, which maintain image prior by attenuating the activation of the temporal attentions, we introduce the layerscale $\mathtt{LS}(\cdot)$ initialized to a very small scalar \citep{layerscale} to the module, referring to the whole layer as the trajectory encoder $\mathtt{TE}(\cdot; \cdot)$,
\begin{align}
\bar{\mathbf{E}}_{l}^{W|j} &:= \mathtt{TE} \bigl( \tilde{\mathbf{E}}_{l}^{W|j}  ;  \tilde{\mathbf{E}}_{l}^{W|\{k<W\}} \bigr) \label{eq:ego_trajectory} \\
&:= \tilde{\mathbf{E}}_{l}^{W|j} +  \mathtt{LS}\bigl( \mathtt{TA}(\tilde{\mathbf{E}}_{l}^{W|j}; \tilde{\mathbf{E}}_{l}^{W|\{k<W\}}) \bigr) \nonumber \\
\bar{\mathbf{T}}_{l}^{W|j} &:= \mathtt{TE} \bigl( \tilde{\mathbf{T}}_{l}^{W|j}  ;  \tilde{\mathbf{T}}_{l}^{W|\{k<W\}} \bigr) \label{eq:target_trajectory} \\
&:= \tilde{\mathbf{T}}_{l}^{W|j} +  \mathtt{LS}\bigl( \mathtt{TA}(\tilde{\mathbf{T}}_{l}^{W|j}; \tilde{\mathbf{T}}_{l}^{W|\{k<W\}}) \bigr). \nonumber
\end{align}

This design ensures that the model is equivalent to the pair-wise model, thus retaining the 3D prior in the initial state.
Throughout the training on dynamic scenes, the model gradually relaxes the degree of attenuation and learns to model complex multi-frame dynamics.

\newpage
\subsection{Feed-forward refinement}
\label{sec:method:refinement}
% Use figure* for multi-column figure
\begin{wrapfigure}{r}{0.5\textwidth}
\vspace{-4mm}
\centering
\includegraphics[width=0.48\textwidth]{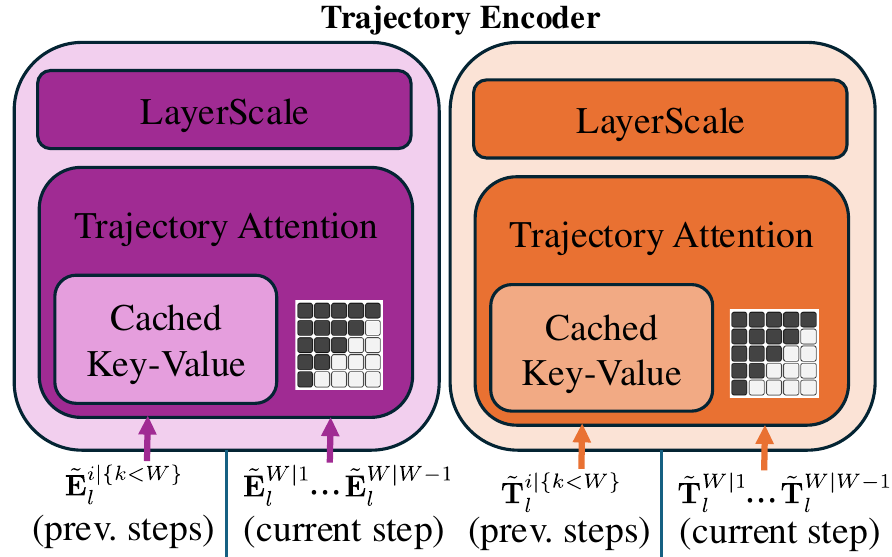}
\caption{\textbf{Illustration of the proposed trajectory encoder.} The trajectory encoder is composed of the trajectory attention with causal masks and the layerscale. The module can refer to the cached key and value tensors, enabling the feed-forward refinement technique.}
\label{fig:ff_refinment}
\end{wrapfigure}

Although~\ours~architecture does not constrain the number of frames $W$,
the finite memory of the system can pose a practical limit.
When processing tens or hundreds of frames as the prediction horizon, a joint processing of whole frames can be impossible.
In order to overcome the limitation, we introduce a feed-forward refinement technique to deal with prediction horizon beyond a chosen $W$.
Specifically, when processing the tokens of an extra frame, \eg$\tilde{\mathbf{E}}_{l}^{i|j}$, where $i\leq W$ and $j > W$, we exploit the pre-computed key and value tensors of $\tilde{\mathbf{E}}_{l}^{i|\{k < W\}}$, which we illustrate in \cref{fig:ff_refinment}.
Since we train the model with the causal attention masking applied to the trajectory attention, these key and value tensors remain equivalent to the case where a larger input size were considered to include the extra frame.
\section{Experiment}
\label{sec:experiment}

In this section, we present the experimental details and compare~\ours~to state-of-the-art baselines. In \cref{sec:experiment:training,sec:experiment:evaluation}, we provide the training details, the data processing, and the evaluation protocols. Then, we experiment, in \cref{sec:experiment:feed_forward}, the feed-forward prediction of pointmaps, and the ablation study in \cref{sec:experiment:ablation}.

\subsection{Training details}
\label{sec:experiment:training}
% \lipsum[14-16]
We initialize the~\ours~model with \duster~\citep{dust3r}, a pair-wise Siamese model pre-trained on scenes covered by 8.5M image pairs from Habitat \citep{habitat}, MegaDepth \citep{megadepth}, StaticThings3D \citep{staticthings}, Apple ARKitScenes \citep{arkit}, BlendedMVS \citep{blendedmvs}, ScanNet \citep{scannet}, Co3D \citep{co3d}, and Waymo \citep{waymo} datasets.
Then, we employ dynamic scenes covered by Point Odyssey \citep{pointodyssey}, Spring \citep{spring}, TartanAir \citep{tartanair}, and Waymo \citep{waymo} datasets to train~\ours~for 4D geometry estimation, follwoing state-of-the-art \monster~\citep{monst3r}.

Despite our design to maintain the strong 3D prior of the pre-trained model \citep{dust3r}, the synthetic scenes in the training dataset can cause a distribution shift in visual texture.
Therefore, we test the trade-off between different training schedules for mixing the synthetic and the real frames, then choose the default setting that demonstrates a balanced performance (see \cref{sec:experiment:ablation} for more details).
Our default setting trains~\ours~for 30 epochs using the AdamW optimizer \citep{adamw} with 20k clips of length $W=5$ per epoch, the mini-batch size 16, and the learning rate $1 \times 10^{-4}$. We sample the clips from real scenes for the first 5 epochs, then employ synthetic scenes for the rest of the training steps. 

\subsection{Evaluation details}
\label{sec:experiment:evaluation}
To evaluate the feed-forward predictions (\cref{sec:experiment:feed_forward}), we employ 3 different test datasets covering dynamic scenes: Point Odyssey \citep{pointodyssey}, Sintel \citep{sintel}, and iPhone dataset \citep{iphone}.
Point Odyssey and Sintel are synthesized scenes generated using 3D rendering engines \citep{pointodyssey, sintel}, and iPhone dataset covers real scenes captured using a synchronized set of camera, lidar, and IMU sensors \citep{iphone}. For each scene, we consider overlapping slices of 12 frames as the evaluation samples.\footnote{We also downsample iPhone dataset \citep{iphone} to 3fps to promote a larger motion.}
We measure the regression accuracy of the pointmaps predicted by~\ours~and the baselines: \duster~\citep{monst3r}, Robust-CVD~\citep{robustcvd}, CasualSAM~\citep{casualsam}, and state-of-the-art \monster~\citep{monst3r}.
Specifically, using a strided sampling, we experiment with $W=2$ (stride 6), $W=4$ (stride 3), and $W=6$ (stride 2) for inference.
As for the metric, we employ the scale and shift invariant error provided by the open source repository of \monster~\citep{monst3r} and report the median error in the target pointclouds per setting: M@2, M@4, and M@6 in \cref{tab:feedforward_prediction}.

\begin{table}[t]
\centering
\resizebox{\columnwidth}{!}{%

\begin{tabular}{l c c c c c c c c c}
\toprule
& \multicolumn{3}{c}{Point Odyssey } & \multicolumn{3}{c}{Sintel} & \multicolumn{3}{c}{iPhone Dataset} \\
\cmidrule(lr){2-4} \cmidrule(lr){5-7} \cmidrule(lr){8-10}
Method & M@2 & M@4 & M@6  & M@2 & M@4 & M@6 & M@2 & M@4 & M@6 \\
\midrule
\duster & 0.547 & 0.549 & 0.552 & 1.595 & 1.865 & 1.598 & \underline{1.301} & \underline{1.532} & \underline{1.716}  \\
Robust-CVD & 0.614 & 0.591 & 0.601 & 1.717 & 1.883 & 1.710 & 1.790 & 1.883 & 2.001 \\
CasualSAM & 0.486 & 0.501 & 0.505 & 1.551 & 1.639 & 1.691 & 1.595 & 1.824 & 1.907 \\
\monster & \underline{0.291} & \underline{0.289} & \underline{0.289} &
\underline{1.374} & \underline{1.411} & \underline{1.433} &
{1.378} & {1.651} & {1.772}  \\
\textbf{\ours} & \textbf{0.264} & \textbf{0.258} & \textbf{0.253} &
\textbf{1.298} & \textbf{1.288} & \textbf{1.287} &
\textbf{1.280} & \textbf{1.436} & \textbf{1.504} \\
\bottomrule
\end{tabular}

}
\caption{
\textbf{Pointmap prediction results.} The quality of pointmaps are compared in terms of the median scale and shift invariant errors with the number of frames 2 (M@2), 4 (M@4), and 6 (M@6). Among the models, \duster~\citep{dust3r}, \monster~\citep{monst3r}, and~\ours~are the feed-forward method, while the others are optimization-based approaches \citep{robustcvd,casualsam}.
}
\label{tab:feedforward_prediction}
\vspace{-4mm}
\end{table}

\subsection{Feed-forward pointmap prediction}
\label{sec:experiment:feed_forward}
\begin{figure}[ht!]
\centering
\includegraphics[width=\linewidth]{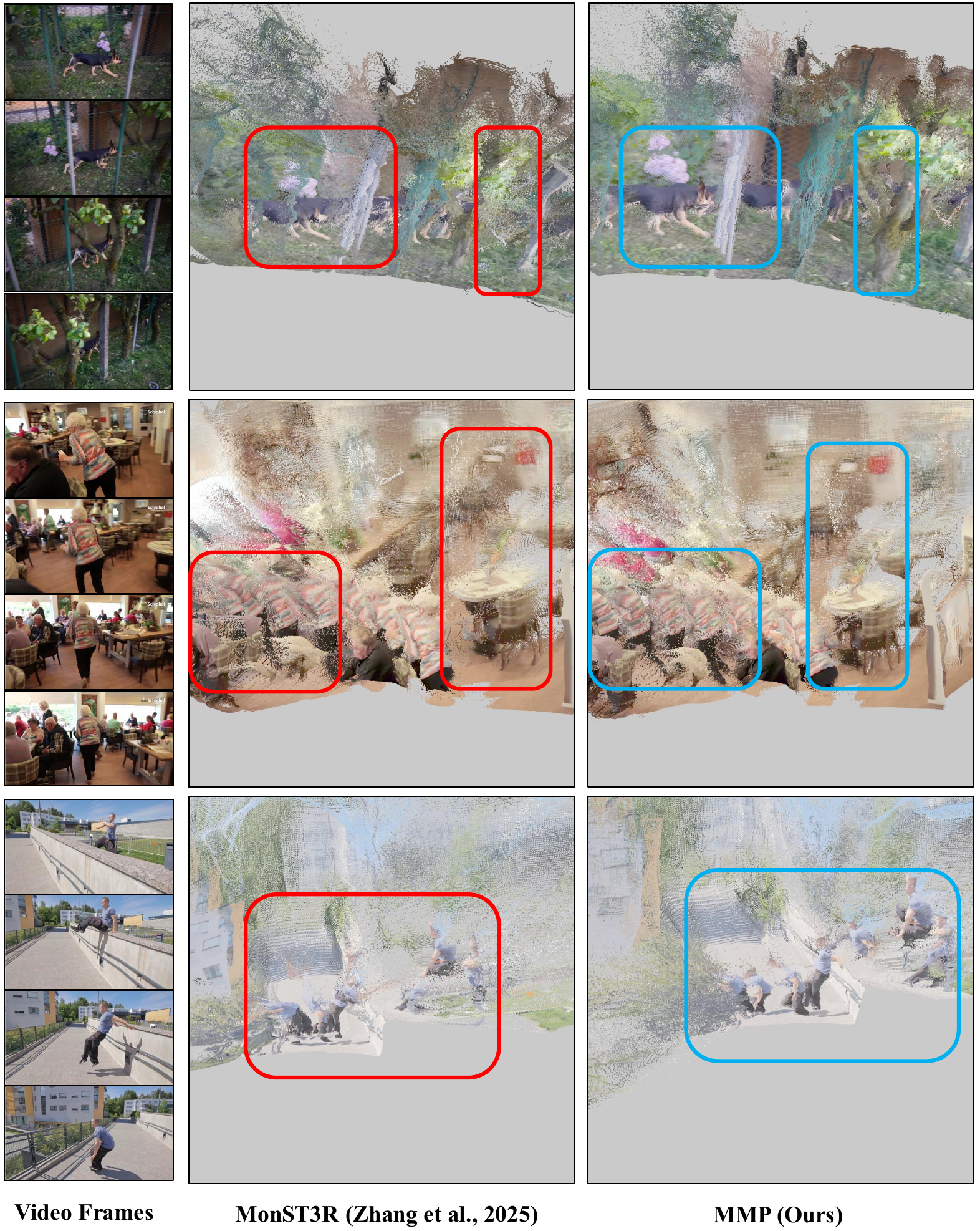}
\caption{\textbf{Qualtative comparison of pointmaps by the baseline \citep{monst3r} and~\ours.} We visualize the the feed-forward pointmaps predicted by \monster~\citep{monst3r} and ours, using video samples from davis dataset \citep{davis}. The inference are performed using $W=8$ frames, where we illustrate even frame indices in the left column.}
\label{fig:qualitative_pointmap}
% \vspace{-6mm}
\end{figure}

In this section, we experiment the feed-forward pointmap prediction by~\ours.
In \cref{tab:feedforward_prediction}, we quantitatively compare the quality of pointmap regression by~\ours~and the baselines: \duster~\citep{monst3r}, Robust-CVD~\citep{robustcvd}, CasualSAM~\citep{casualsam}, and \monster~\citep{monst3r}.
Next, we provide the visualization of the pointmaps produced by~\ours~in \cref{fig:qualitative_pointmap}, executed on DAVIS video frames \citep{davis}.

To begin with, we find~\ours~can outperform the strongest feed-forward baseline, \monster~\citep{monst3r}, \eg 15.1\% improvement M@6 1.772 (\monster~\citep{monst3r}) $\to$ 1.504 (\ours) on iPhone dataset \citep{iphone} in \cref{tab:feedforward_prediction}. 
While our method is trained on the same data distribution as the baseline, an enhanced performance is observed even under a pair-wise inference (\ie M@2). This supports the significance of the trajectory encoder employed in our method, which facilitates learning useful representation for predicting accurate pointmaps.
\ours~can consistently improve the quality of dynamic pointmaps compared to the baselines in various scenarios covering synthetic and real video scenes.
We also note that our method can demonstrate the results that are more robust over various strides, (\eg comparing to \monster~\citep{monst3r} in Sintel \citep{sintel}: 1.374 $\to$ 1.298 M@2, 1.411 $\to$ 1.288 M@4, and 1.433 $\to$ 1.287 M@6), which we attribute to the dynamics modeling enabled by our method.

% Use figure* for multi-column figure

From the qualitative study in \cref{fig:qualitative_pointmap}, we find our method tends to demonstrate more accurate pointmaps over the frames, \eg the background objects and the scene are consistently depicted, comparing the regions indicated by red boxes (\monster~\citep{monst3r}) and the blue boxes (\ours), which reveals the efficacy of our method in complex dynamic scenes.

\subsection{Ablation study}
\label{sec:experiment:ablation}
In this section, we perform ablation study for the effect of proposed techniques in this paper, namely the trajectory encoder and the scheduled training, and compare different training schedules in terms of the average of $\{\text{M@2},\text{M@4},\text{M@6}\}$.
In \cref{tab:module_ablation}, we find employing the trajectory encoder is indeed significant to the performance of~\ours, and the scheduled training can mitigate the negative effect of the synthetic training data on the performance.

We further study the effect of applying different schedules for training synthetic and real scenes by~\ours~in \cref{tab:schedule_ablation}, which
compares 4 different training strategies: synthetic only (\ie synthetic scenes for 30 epochs), joint training (\ie mixed data for 30 epochs), synthetic then real (\ie synthetic scenes for the first 25 epochs, then real scenes for the rest 5 epochs), and real then synthetic (the default setting).
While there exist trade-offs in the performances over the datasets, we choose the real then synthetic schedule as our final design, which can demonstrate a balanced performance.

\begin{table}[t]
\centering
\begin{tabular}{l c c c}
\toprule
Model & Point Odyssey  & Sintel & iPhone Dataset \\
\midrule
Vanilla Siamese & 0.290 & 1.406 & 1.600 \\
+ Trajectory Encoder & \textbf{0.237} & \textbf{1.011} & \underline{1.571} \\
+ Scheduled Training & \underline{0.258} & \underline{1.291} & \textbf{1.407} \\
% + Joint Static Training & - & - & - \\
\bottomrule
\end{tabular}
\caption{
\textbf{Ablation study.} The effect of trajectory encoder and the scheduled training is studied in terms of the average pointmap regression errors.
}
\label{tab:module_ablation}
\vspace{-2mm}
\end{table}

\begin{table}[t]
\centering
\begin{tabular}{l c c c}
\toprule
Model & Point Odyssey  & Sintel & iPhone Dataset \\
\midrule
Synthetic Only & \textbf{0.237} & \textbf{1.011} & 1.571 \\
Joint Training & 0.266 & 1.393 & 1.439 \\
Synthetic then Real & 0.271 & 1.440 & \textbf{1.383} \\
Real then Synthetic & \underline{0.258} & \underline{1.291} & \underline{1.407} \\
\bottomrule
\end{tabular}
\caption{
\textbf{Comparison of training schedules.} The effect of training schedules is studied in terms of the average pointmap regression errors.
}
\label{tab:schedule_ablation}
\vspace{-5mm}
\end{table}

\section{Discussion}
\label{sec:discussion}
In this section, we discuss the extreme cases in relation to the fundamental assumption considered by~\ours~and the pair-wise baseline \citep{monst3r}. Next, we further discuss the limitation of~\ours~and future research directions.

\subsection{Extreme case}
\label{sec:discussion:extream_case}
Although the pair-wise architecture \citep{dust3r, monst3r} can produce pointmaps for more than 2 frames by executing multiple pair-wise inferences, its design inevitably enforces the assumption that the distributions of consecutive pointmaps are independent. For example, given $\{\mathbf{I}^i, \mathbf{I}^j, \mathbf{I}^k\}$, a pair-wise model assumes that a joint density $\mathtt{Pr}(\mathbf{Y}^{i|j},\mathbf{Y}^{i|k},\mathbf{Y}^{j|k})$ is proportional to $\mathtt{Pr}(\mathbf{Y}^{i|j})\cdot \mathtt{Pr}(\mathbf{Y}^{i|k}) \cdot \mathtt{Pr}(\mathbf{Y}^{j|k})$.

However, in practice, including the scenarios represented by our evaluation, there exists an extreme case where $\mathbf{I}^i$ and $\mathbf{I}^k$ are completely non-overlapping, so that the pair-wise model assigns an erroneous estimate of $\mathtt{Pr}(\mathbf{Y}^{i|k})$, which can induce significant failure modes of estimating the joint density.
Even if the global optimization is employed, depending on the sampling strategy, there is a potential extreme case that the connectivity becomes independent. To prevent the case, a sophisticated hyperparameter engineering would be required.
Since~\ours~can relax this constraint up to $W$ frames and beyond (with the feed-forward refinement), it can learn the pointmap distribution that is more close to the true nature of the dynamic scenes.
For example, the intriguing tendency of~\ours~in \cref{tab:feedforward_prediction}, being robust to the evaluation stride can be attributed to a more accurate estimation of the joint density over a set of frames.

\subsection{Limitation}
\label{sec:discussion:limitation}
Despite the promising results demonstrated by~\ours, the scarcity of 4D dyanamic scenes can hinder the generalization performance.
To mitigate the distribution shifts, we employ the scheduled training to maintain the visual texture prior in the pre-trained model.
However, since we still observe trade-offs in the performance, as shown in \cref{tab:schedule_ablation},
designing new training datasets, self-supervised learning with unlabeled data, or an objective functions robust to the distribution shift for 4D geometry estimation can be interesting future directions.
It is also worth noting that we focus on the realistic scenarios where the observation is captured by a monocular video camera, rather than multiple synchronized cameras capturing one scene.
Although it would be straightforward to apply~\ours~for the synchronized cameras, we believe that there is a room to exploit useful properties, such as epipolar geometry \citep{hartley2003multiple} of the synchronized cameras, which is another interesting future direction.

\section{Conclusion}
\label{sec:conclusion}
In this paper, we propose~\ours, a feed-forward 4D geometry estimation model for dynamic pointmaps.
We tackle the limitation in existing baselines based on the pair-wise Siamese architecture, being sub-optimal under complex dynamic scenes.
For example, we propose to encode point-wise dynamics on the pointmap representation for each frame, enabling significantly improved expressiveness for dynamic scenes.
In the experiments, we find our method can outperform the state-of-the-art in terms of the regression accuracy of the feed-forward prediction.

\textbf{Acknowledgements}. This paper was supported by RLWRLD.

{\small
\bibliography{11_references}
\bibliographystyle{unsrtnat}
}

\end{document}